\documentclass[sigconf]{acmart}
\usepackage{enumitem}
\usepackage{tabularx}
\usepackage{graphicx}
\usepackage{caption}
\usepackage{array}
\usepackage{pifont}

\acmConference[OARS @ RecSys'26]{ACM RecSys Workshop on Online \& Adaptive Recommender System}{September 28, 2026}{Minneapolis, MN, USA}
\acmDOI{}
\acmISBN{}
\allowdisplaybreaks

\begin{document}

\title{Next Best Action: Causal Constrained 
Offline RL for Business Action Targeting}

\setcopyright{none}
\renewcommand\footnotetextcopyrightpermission[1]{}

\author{Surya Ramachandiran}
\affiliation{%
  \institution{Lyft, Inc.}
  \city{San Francisco}
  \state{CA}
  \country{USA}}
\email{surya.ramachandiran18@gmail.com}

\author{Yiming Li}
\affiliation{%
  \institution{Lyft, Inc.}
  \city{San Francisco}
  \state{CA}
  \country{USA}}
\email{yli@lyft.com}

\author{Juan Mancilla-Caceres}
\affiliation{%
  \institution{Lyft, Inc.}
  \city{San Francisco}
  \state{CA}
  \country{USA}}
\email{juanmc@lyft.com}

\author{Tao Ye}
\affiliation{%
  \institution{Lyft, Inc.}
  \city{San Francisco}
  \state{CA}
  \country{USA}}
\email{taoye@lyft.com}

\author{Jun Liu}
\affiliation{%
  \institution{Lyft, Inc.}
  \city{San Francisco}
  \state{CA}
  \country{USA}}
\email{junliu@lyft.com}

\begin{abstract}
Next Best Action (NBA) systems must select among customer interventions (e.g., discounts, promotions, and lifecycle actions), maximize long-term economic value, satisfy complex operational constraints, and remain causally grounded. Existing approaches lack a rigorous framework for meeting these requirements jointly. We address this gap by formulating NBA as a Constrained Markov Decision Process (CMDP). We then integrate off-policy causal correction of the estimated customer dynamics directly into the Linear Programming (LP) formulation of the CMDP. The resulting policy selects actions based on customer states to maximize long-term value while explicitly enforcing business guardrails. Furthermore, the LP formulation yields shadow prices that quantify the economic trade-offs of operational constraints (e.g., fairness and user welfare), enhancing policy interpretability. We validate our proposed offline Reinforcement Learning (RL) framework on an industry case study involving rideshare coupon targeting, where customer states are defined by Recency, Frequency, and Monetary (RFM) stages. We evaluate performance using model-based off-policy evaluation, with causal dynamics estimated from a temporal holdout. Under the observed coupon budget and RFM state representation, our optimized policy improves customer lifetime value (LTV) by 19\% over the estimated behavior policy and 8\% over a myopic baseline. Although the demonstrated action space is binary, the proposed framework is action-agnostic, yielding a general-purpose decision module for targeting systems that manage multiple interventions.

\end{abstract}



\maketitle

\section{Introduction}

\begin{figure}[h]
    \centering
    \includegraphics[width=0.45\textwidth]{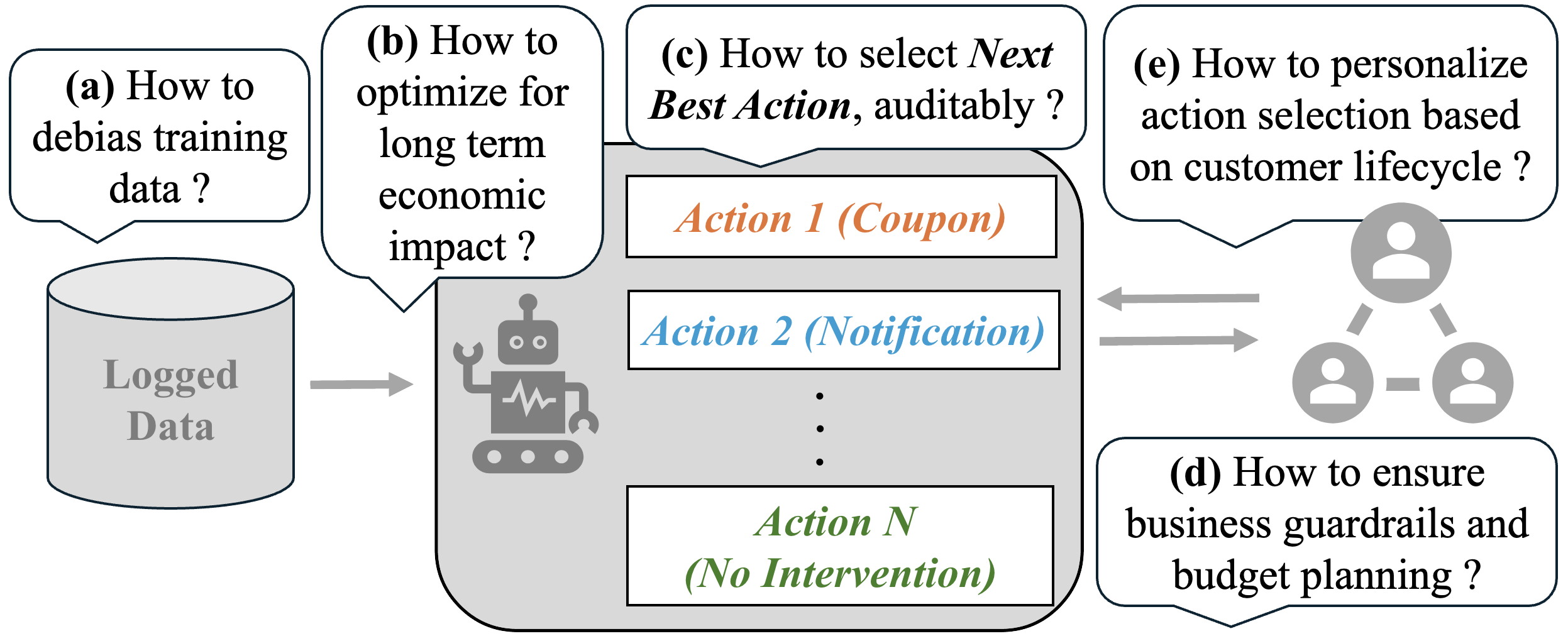}
    \Description{Challenges for business action targeting systems}
    \captionsetup{skip=2pt}
    \caption{Challenges for business action targeting systems}
    \label{fig:challenges}
\end{figure}

Consumer platforms increasingly employ \emph{business action targeting} systems to determine the optimal timing and type of intervention (e.g., discounts, promotions, or lifecycle actions) for each customer to maximize long-term engagement \cite{zheng2018drn, zou2019reinforcement, ie2019slateq, chen2019top, xu2023optimizing}. Developing such systems raises several coupled challenges illustrated in Figure~\ref{fig:challenges}, yet most existing research addresses them in isolation.

The first challenge is \emph{training data bias}. Because logged interactions are generated by a historical targeting policy rather than a randomized experiment, naive estimates of action effects are heavily confounded \cite{bottou2013counterfactual, schnabel2016recommendations}. To correct for this, large-scale recommenders commonly employ off-policy estimators such as propensity-based, doubly robust, and self-normalized importance sampling methods \cite{rosenbaum1983central, dudik2011doubly, swaminathan2015counterfactual, swaminathan2015self, chen2019top, jeunen2021pessimistic, gupta2024optimal}. These approaches, however, predominantly target single-step (bandit) outcomes rather than multi-step customer dynamics.

A second theme of research addresses this gap by framing recommendation and targeting as sequential decision-making problems. Early MDP recommenders \cite{shani2005mdp} and lifetime value targeting models \cite{theocharous2015personalized} have since matured into industrial deep RL systems that optimize long-term engagement \cite{zheng2018drn, zou2019reinforcement, ie2019slateq, chen2019top, xu2023optimizing}. Typically trained offline \cite{levine2020offline}, these methods rely on pessimism \cite{kumar2020conservative} or stationary distribution corrections (e.g., DICE frameworks \cite{nachum2019dualdice, lee2022coptidice}) to mitigate out-of-distribution extrapolation error. However, standard offline RL rarely guarantees compliance with complex business constraints (e.g., budget caps, exposure limits, fairness), while DICE methods leave observational confounding unaddressed and remain unstudied in recommendation and targeting contexts.

A third theme, popular in both research and industry, tackles business constraints directly by casting targeting as a two-step incentive allocation problem that pairs an uplift model \cite{gutierrez2017causal} with mathematical optimization \cite{goldenberg2020free, zhao2019unified, albert2022ecommerce, zhou2023direct}. While effective at enforcing budgets and business guardrails, this approach is myopic. It maximizes immediate conversion or short-term uplift while ignoring how current actions shape long-term customer value.

In short, each existing approach addresses only a subset of the challenges shown in Figure~\ref{fig:challenges} (e.g., correcting for causal bias without optimizing for the long horizon, maximizing long-term value without exact constraints or business guardrails, or allocating under constraints without long-term objectives). To our knowledge, this is the first framework that unifies causal off-policy correction with constrained RL to enable auditable, constraint-verifiable business intervention targeting in an industry setting.

\section{Methodology}

\begin{figure}[htbp]
    \centering
    \includegraphics[width=0.45\textwidth]{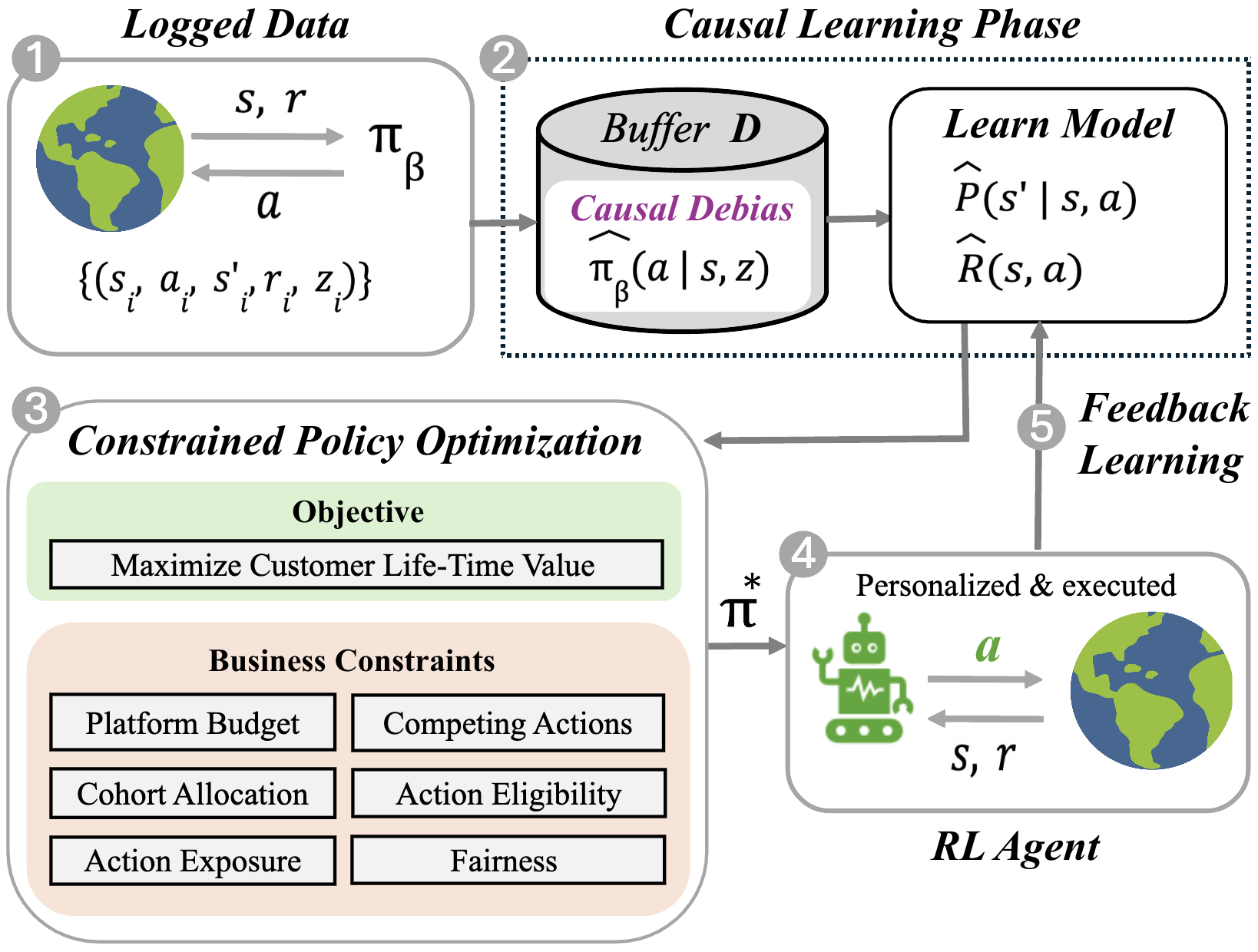}
    \Description{Next Best Action (NBA) framework}
    \captionsetup{skip=2pt}
    \caption{Next Best Action (NBA) framework}
    \label{fig:overview}
\end{figure}

Our Next Best Action (NBA) framework (see Figure~\ref{fig:overview}) addresses these challenges (see Figure~\ref{fig:challenges}): (a) because the logged training data reflect a biased prior policy, we propose to use  propensity weighting \cite{swaminathan2015self, rosenbaum1983central} in model-based offline RL \cite{levine2020offline} to causally correct the learned transition and reward estimates; (b) leveraging these corrected estimates, we formulate action selection as a CMDP to directly optimize for long-term value rather than immediate rewards or clicks; (c) we solve the CMDP using its dual Linear Program (LP), which allows us to choose the right next action verifiably \cite{altman2021constrained, puterman2014markov}; (d) the LP formulation allows us to model business guardrails (like budget caps, exposure limits, and fairness) as explicit constraints, while also allowing for budget planning and scenario analysis; (e) the resulting optimal policy is explainable as the decisions are adapted to the user lifecycle stages \cite{chen2012data}. Because every action carries a transparent, causal, and constraint-verifiable rationale, the system is auditable by design. 

While highly interpretable, the current formulation is restricted to discrete states, leaving continuous extensions for future work. Offline RL~\cite{levine2020offline} is also vulnerable to extrapolation errors caused by limited data coverage. Because the initial case study uses discrete user lifecycle states, every state-action pair is well-supported by the industrial dataset. This strong coverage mitigates the out-of-distribution extrapolation errors~\cite{kumar2020conservative, fujimoto2019off} that typically plague offline RL. However, as we generalize to richer states and several actions, data coverage becomes sparse, requiring pessimism to be modeled explicitly \cite{kumar2020conservative, levine2020offline, jeunen2021pessimistic}.

\subsection{Constrained  RL Problem Notation}

We model user-agent interactions as a CMDP $\langle \mathcal{S}, \mathcal{A}, P, R, \gamma, \mu, \omega, B \rangle$ \cite{altman2021constrained, puterman2014markov}. Here $\mathcal{S}$ is the set of discrete customer lifecycle states and $\mathcal{A}$ is the set of actions, where $a_0 \in \mathcal{A}$ denotes the \emph{No Intervention} action and $\mathcal{A}_{\mathrm{I}} = \mathcal{A} \setminus \{a_0\}$ is the set of \emph{Intervention} actions. The dynamics are given by the transition probabilities $P(s' \mid s, a)$ and the immediate rewards $R(s, a)$. The time discount factor is denoted by $\gamma \in (0, 1)$, and initial state distribution is given by $\mu(s)$. To enforce business guardrails, $\omega \in \mathbb{R}^{|\mathcal{A}|}$ specifies the state-invariant action costs, with $\omega(a_0) = 0$. Finally, $B \in \mathbb{R}$ is the scalar platform budget cap. The remaining operational thresholds (cohort, exposure, competition, and fairness) are introduced alongside their respective constraints.

\subsection{Learn Causal Model Dynamics}

To account for logging policy bias, we first estimate the propensity scores $\hat\pi_b(a \mid s,z)$~\cite{rosenbaum1983central} via multinomial logistic regression on the historical data, conditioning on observed states $s$ and confounders $z$ that span historical behavior and context. Using $\hat\pi_b$, we compute Inverse Propensity Weights (IPW) $\hat{w}_i = 1/\hat\pi_b(a_i \mid s_i, z_i)$ for each observed one-step transition $i$. To mitigate high variance associated with the standard IPW, we use Self-Normalized Inverse Propensity Weights (SNIPW) and estimate causally corrected transition probabilities $\hat P(s'\mid s,a)$ and rewards $\hat R(s,a) $ as shown in Equation~\eqref{eq:causal_eqn}.

\begin{equation}\label{eq:causal_eqn}
\hat P(s'\mid s,a) = \frac{\displaystyle\sum_{i\in\mathcal{D}_{s,a,s'}} \hat w_i}{\displaystyle\sum_{i\in\mathcal{D}_{s,a}} \hat w_i},
\qquad \hat R(s,a) = \frac{\displaystyle\sum_{i\in\mathcal{D}_{s,a}} \hat w_i\, r_i}{\displaystyle\sum_{i\in\mathcal{D}_{s,a}} \hat w_i}
\end{equation}

The causal validity of these re-weighted estimates rests on the \emph{strong ignorability}~\cite{rosenbaum1983central} assumption. This comprises two conditions: \emph{unconfoundedness}, meaning $(s,z)$ captures all common causes of the action and outcomes, and \emph{positivity}, meaning $\pi_b(a \mid s,z) > 0$ for every action $a$ available in state $s$ and all $(s,z)$ in the data. Under this assumption, the re-weighted estimators $\hat{P}(s'\mid s,a)$ and $\hat{R}(s,a)$ recover the true interventional transition distribution $P(s'\mid\text{do}(a), s)$ and the true interventional reward $R(s, \text{do}(a))$, respectively, rather than their confounded observational counterparts. Any unobserved confounding violates this assumption, introducing residual bias in $\hat{P}$ and $\hat{R}$ that propensity weighting alone cannot eliminate. To mitigate any  numerical instability that may arise from extreme propensity weights, we consider percentile-based clipping to the weights.

\subsection{Linear Program}
The CMDP is formulated through its dual LP using the state-action occupancy measure ($\zeta(s,a)$) \cite{altman2021constrained, puterman2014markov}. The formulation maximizes the expected discounted cumulative reward, subject to model dynamics (see Equation~\eqref{eq:causal_eqn}) and comprehensive business constraints (see Table~\ref{tab:business_constraints}). The optimal state-action occupancy measure $\zeta^\star(s,a)$, obtained from solving the LP, is used to extract the optimal policy $\pi^\star(a \mid s)$ \cite{puterman2014markov}.

\begin{table}[h]
\small
\caption{Supported classes of business constraints}
\label{tab:business_constraints}
\begin{tabularx}{\columnwidth}{>{\raggedright\arraybackslash\bfseries}p{2.8cm}X}
\toprule
Constraint Class & Operational Definition / Business Goal \\
\midrule
Platform Budget & Caps overall system level incentive/spend. \\
Cohort Allocation & Restricts or guarantees intervention spend within target segments. \\
Action Exposure & Caps frequency per intervention type to prevent user fatigue. \\
Competing Actions & Allocates a shared, finite budget across mutually exclusive actions.\\
Action Eligibility & Excludes invalid state-action pairs (e.g., subscription targeting for already subscribed). \\
Fairness Guardrails & Guarantees minimum action exposure to prevent any systematic segment neglect. \\
\bottomrule
\end{tabularx}
\end{table}

\begin{align}
\max_{\zeta} \quad
  & \sum_{s \in \mathcal{S},\, a \in \mathcal{A}} \zeta(s,a)\, \hat{R}(s,a)
    \label{eq:objective} \\
\text{s.t.} \quad
  & \sum_{a \in \mathcal{A}} \zeta(s,a) \notag \\
  & \qquad - \gamma \sum_{\substack{s' \in \mathcal{S} \\ a' \in \mathcal{A}}}
      \hat{P}(s \mid s', a')\, \zeta(s', a')
    = \mu(s),
    && \forall s \in \mathcal{S}
    \label{eq:flow_conservation} \\[3pt]
  & (1-\gamma) \sum_{s \in \mathcal{S},\, a \in \mathcal{A}} \omega(a)\, \zeta(s,a) \le B,
    \label{eq:budget} \\[3pt]
  & \underline{B}_g \le (1-\gamma) \sum_{s \in \mathcal{S}_g,\, a \in \mathcal{A}} \omega(a)\, \zeta(s,a) \le \overline{B}_g,
    && \forall g \in \mathcal{G}
    \label{eq:cohort} \\[3pt]
  & (1-\gamma) \sum_{s \in \mathcal{S}} \zeta(s,a) \le E_a,
    && \forall a \in \mathcal{A}_{\mathrm{I}}
    \label{eq:exposure} \\[3pt]
  & (1-\gamma) \sum_{s \in \mathcal{S},\, a \in \mathcal{A}_j} \omega(a)\, \zeta(s,a) \le C_j,
    && \forall j \in \mathcal{J}
    \label{eq:competing} \\[3pt]
  & \zeta(s,a) = 0,
    && \forall (s,a) \in \mathcal{E}
    \label{eq:eligibility} \\[3pt]
  & (1-\gamma) \sum_{s \in \mathcal{S}_g} \zeta(s,a) \ge F_{g,a},
    && \forall g \in \mathcal{G},\, a \in \mathcal{A}_{\mathrm{I}}
    \label{eq:fairness} \\[3pt]
  & \zeta(s,a) \ge 0,
    && \forall s \in \mathcal{S},\, a \in \mathcal{A}
    \label{eq:nonneg}
\end{align}

The objective~\eqref{eq:objective} maximizes the expected discounted cumulative reward, which is linear in the occupancy measure $\zeta(s,a)$. Here $\zeta(s,a)$ denotes the total expected discounted number of times the agent visits state $s$ and takes action $a$ under a given policy. Because $\zeta$ is unnormalized, every rate constraint carries a $(1-\gamma)$ factor that converts occupancy mass into a normalized frequency~\cite{altman2021constrained}.

The constraints jointly define the feasible policy space. Equation~\eqref{eq:flow_conservation}
imposes \emph{Bellman flow conservation} under the causally corrected dynamics $\hat{P}$, ensuring that $\zeta$ is a valid occupancy measure; the remaining constraints encode the business guardrails and operational constraints mentioned in Table~\ref{tab:business_constraints}. Specifically, Equation~\eqref{eq:budget}
caps the system-level intervention cost at $B$, while
Equation~\eqref{eq:cohort} confines per cohort spend to
$[\underline{B}_g,\overline{B}_g]$. Equation~\eqref{eq:exposure} bounds each intervention's long-run frequency (e.g., $E_a=0.05$ limits exposure to promotions to at most $5\%$ of interactions).
Equation~\eqref{eq:competing} caps the shared budget of mutually exclusive and competing action groups. Equation~\eqref{eq:eligibility} prohibits ineligible state-action decisions. Equation~\eqref{eq:fairness} guarantees a minimum per cohort exposure to various interventions to ensure fairness. Finally, Equation~\eqref{eq:nonneg} enforces non-negativity so that $\zeta$ remains a valid measure. 

On solving the above LP, we directly obtain the optimal state-action occupancy measure $\zeta^\star(s,a)$. The corresponding optimal policy $\pi^\star(a \mid s)$ is then extracted by normalizing over actions (see Equation \eqref{eq:policy_extraction}). For states with $\sum_{a} \zeta^\star(s,a) = 0$, the normalization is undefined; such states carry zero occupancy mass, and to preserve the budget constraint under a plausible distribution shift at deployment, we default to the \emph{No Intervention} action ($\pi^\star(a_0 \mid s) = 1$ and $\pi^\star(a \mid s) = 0$ for all $a \in \mathcal{A}_{\mathrm{I}}$). In rare cases, the LP solution may have multiple optima $\zeta^\star$ (e.g., two actions with identical long-term value at a state); the LP solver breaks such ties arbitrarily, with no effect on the objective. Finally, because the problem is constrained, the resulting optimal policy is in general \emph{stochastic}. 

\begin{equation}
\pi^\star(a \mid s) = \frac{\zeta^\star(s,a)}{\sum_{a' \in \mathcal{A}} \zeta^\star(s,a')}.
\label{eq:policy_extraction}
\end{equation}

Solving the LP also provides us with shadow prices associated with each constraint, which serve as an auditable rationale for the resulting allocation and support budget planning. In particular, the dual variable $\lambda_B^\star$ of the platform-level budget constraint (Equation~\eqref{eq:budget}) gives the marginal gain in expected long-term value per additional unit of spend, i.e., the return on the next budget dollar. Similarly, the duals of the cohort, exposure, competing actions, and fairness constraints are the shadow prices of the respective business guardrails. Each provides the marginal long-term value forgone by enforcing that guardrail, or equivalently, the value to gain from relaxing it by one unit. Finally, nonzero duals identify \emph{binding} constraints actively shaping the policy, while zero duals indicate \emph{slack}, making the trade-offs of every decision transparent and verifiable.

\section{Case Study}

To evaluate the NBA framework, we conduct an offline coupon-targeting case study using 68 weeks of proprietary historical data (consisting of a 52 week training period and a subsequent 16 week temporal holdout) from the rideshare platform with millions of active users. We map historical weekly rider interactions into discrete states, actions, rewards, and observed confounders. States are defined via RFM (Recency, Frequency, and Monetary) segmentation \cite{chen2012data}, classifying customers into eight lifecycle stages (e.g., \textit{Champion}, \textit{Hibernating}) based on the RFM values of their rides in the past 28 days (see Table~\ref{tab:rfm_states}). The \textit{Coupon Action} is a ride discount offered to a customer in a given week, incentivizing them to ride; \textit{No Action} does not provide this incentive. The immediate reward is a blended business value metric generated by that customer in that week. NBA maximizes the expected discounted cumulative reward a customer generates, which maps to customer LTV \cite{berger1998customer}, i.e., the long-run business value of a customer. The observed confounders are comprehensive and span historical behavior and ride context.

Following extant literature \cite{pfeifer2000modeling, chen2012data}, we assume RFM-based representations approximately satisfy the Markov property. We also independently validated that RFM-based features are the strongest predictors of rider LTV. Furthermore, given the few, finite states ($|\mathcal{S}| = 8$) and binary action ($|\mathcal{A}| = 2$), our industrial dataset provides high data support across all 16 state-action pairs.

We benchmark candidate policies using model-based off-policy evaluation on a temporal holdout. The optimal policy $\pi^\star$ is trained on the first 52 weeks of data ($D_{\text{train}}$) and evaluated on SNIPW-corrected transition dynamics $\hat{P}_{\text{eval}}$ and reward dynamics $\hat{R}_{\text{eval}}$, estimated from the subsequent 16 week holdout ($D_{\text{eval}}$). To prevent instability in the SNIPW estimator, we clip propensity weights at the 99th percentile, following standard practice in causal inference.

\begin{table}[htbp]
\centering\small
\caption{Case study formulation}
\label{tab:problem-formulation}
\begin{tabular}{@{}ll@{}}
\toprule
Components & Description \\
\midrule
Learning Episode ($\tau$)   & Sequence of a rider's states, actions, rewards, \\
                            & and observed confounders over 52 weeks \\ 
Time step ($t$)             & Weekly decision stages \\
States ($s_t \in S$)        & Rider lifecycle stages, based on RFM \\
Action ($a_t \in A$)        & (1) No Action (2) Coupon Action \\
Rewards ($r_t$)             & Blended business metric   \\
Discount factor ($\gamma$)      & Weights on future rewards, $\gamma = 0.9$ \\
Optimal Policy ($\pi^\star$)    & $\arg\max_\pi \mathbb{E}_\pi \left[ \sum_{t=1}^{\infty} \gamma^{t-1} R(s_t, a_t) \right]$ \\
                            & subject to constraints (i.e. maximizes \\
                            & constrained rider LTV) \\
\bottomrule
\end{tabular}
\end{table}

\begin{table}[h]
  \caption{RFM segmentation states (H=High, L=Low)}
  \label{tab:rfm_states}
  \centering\small
  \begin{tabular}{lcccccccc}
    \toprule
    States & $s_1$ & $s_2$ & $s_3$ & $s_4$ & $s_5$ & $s_6$ & $s_7$ & $s_8$ \\
    \midrule
    Narrative&
    \rotatebox{75}{Champion} &
    \rotatebox{75}{Loyalist} &
    \rotatebox{75}{Promising} &
    \rotatebox{75}{New} &
    \rotatebox{75}{At Risk} &
    \rotatebox{75}{Slipping} &
    \rotatebox{75}{Hibernating} &
    \rotatebox{75}{Dormant} \\[2ex]
    Recency   & H & H & H & H & L & L & L & L \\
    Frequency & H & H & L & L & H & H & L & L \\
    Monetary  & H & L & H & L & H & L & H & L \\
    \bottomrule
  \end{tabular}
  \\ \raggedright \small \textit{Note: The High/Low classification is determined by a static median split.}
\end{table}

\section{Results and Discussion}
In this section, we analyze the estimated causal transition and reward dynamics, benchmark the obtained optimal policy in terms of long-term value, and examine the optimal coupon policy and its returns across varying budget constraints.

\subsection{Causal Transition and Reward Dynamics}

\begin{figure}[h]
    \centering
    \includegraphics[width=0.45\textwidth]{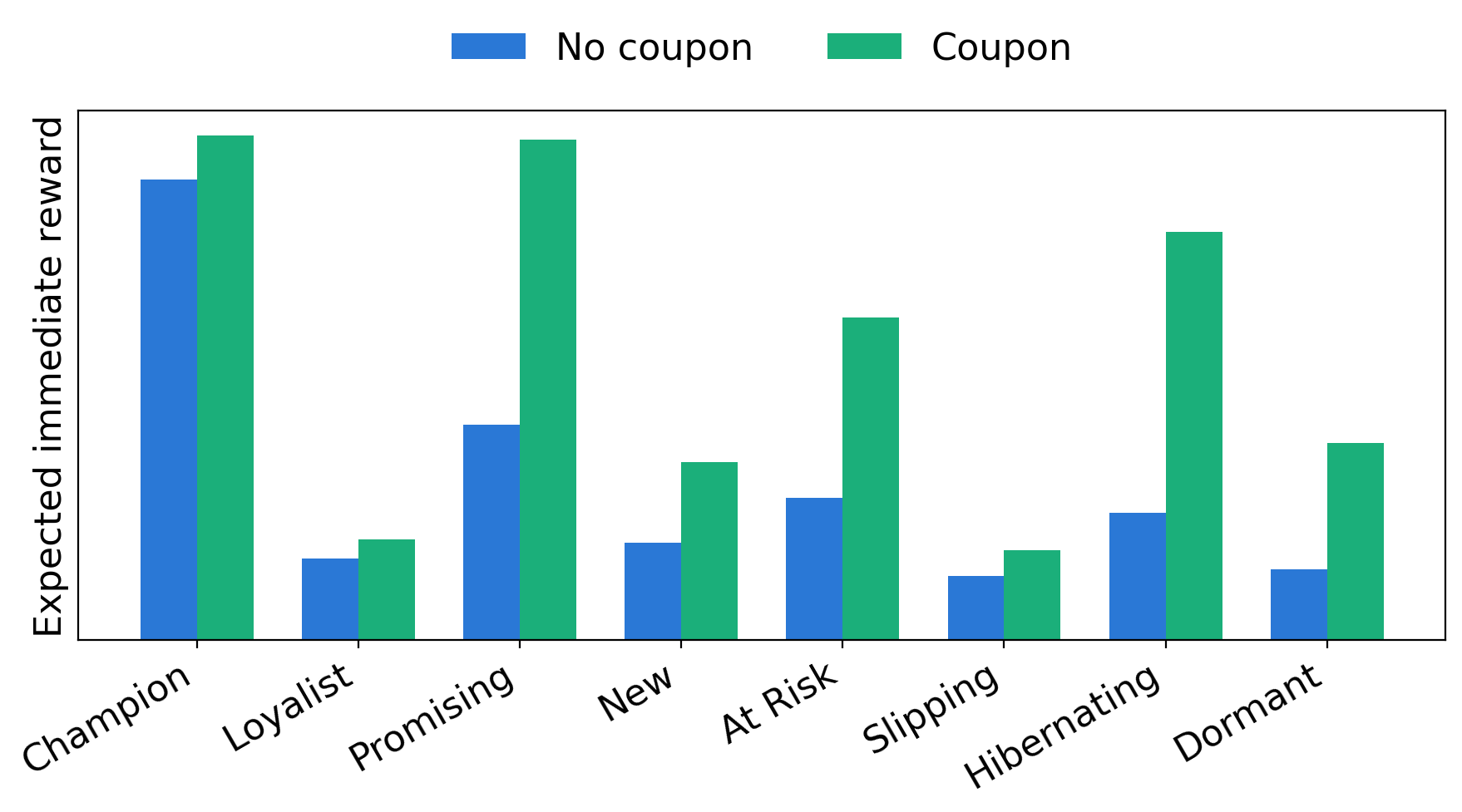}
    \Description{Learned causal reward estimates}
    \captionsetup{skip=2pt}
    \caption{Learned causal reward estimates}
    \label{fig:rewards}
\end{figure}

\begin{figure}[h]
    \centering
    \includegraphics[width=0.45\textwidth]{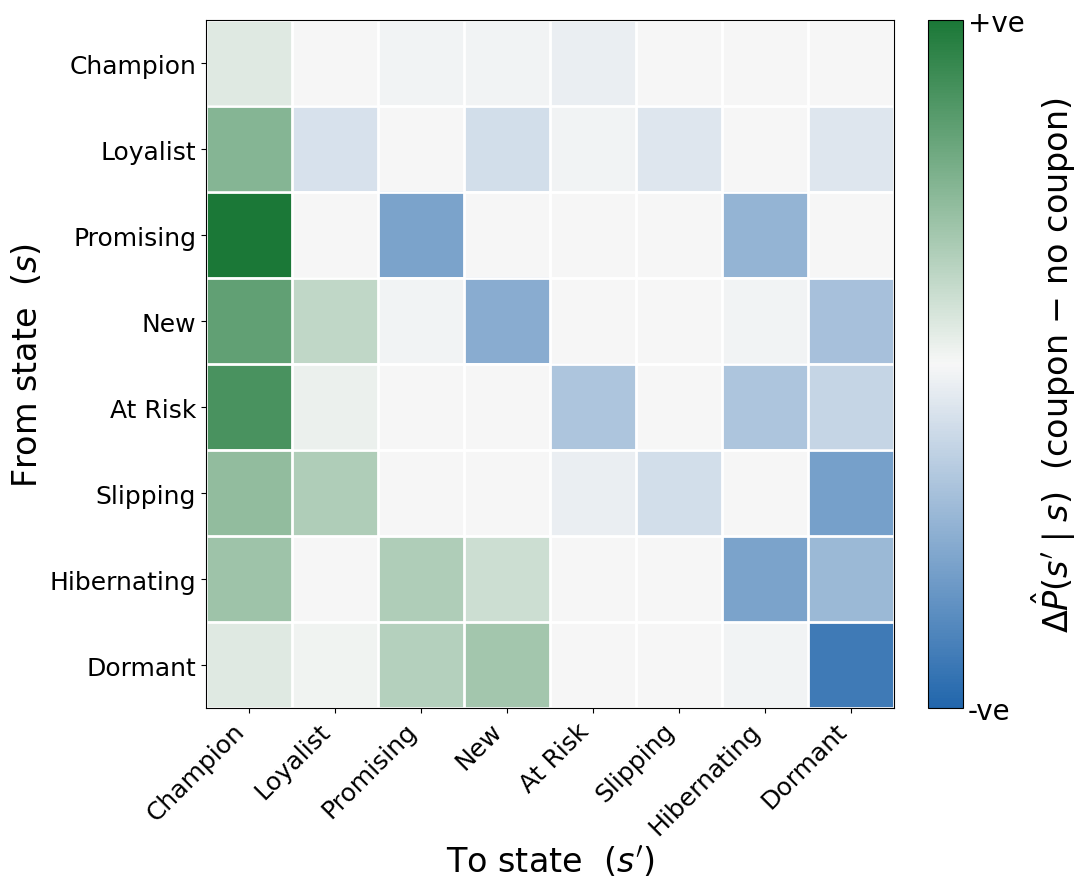}
    \Description{Causal effect of coupon on state transitions}
    \captionsetup{skip=2pt}
    \caption{Causal effect of coupon on state transitions}
    \label{fig:state_tran}
\end{figure}

Figure~\ref{fig:rewards} presents the expected immediate rewards $\hat{R}(s, a)$ after causal correction. Under the coupon action, the \textit{Hibernating} segment yields nearly three times its baseline reward, while the \textit{Champion} segment shows only a marginal gain. Similarly, Figure~\ref{fig:state_tran} illustrates the causal effect of the coupon action on state transitions $\hat{P}(s' \mid s, a)$. Exposing the \textit{Promising} segment to a coupon more than doubles its probability of transitioning to the \textit{Champion} state relative to the baseline.

\subsection{Policy Performance Benchmarking}

\begin{figure}[h]
    \centering
    \includegraphics[width=0.45\textwidth]{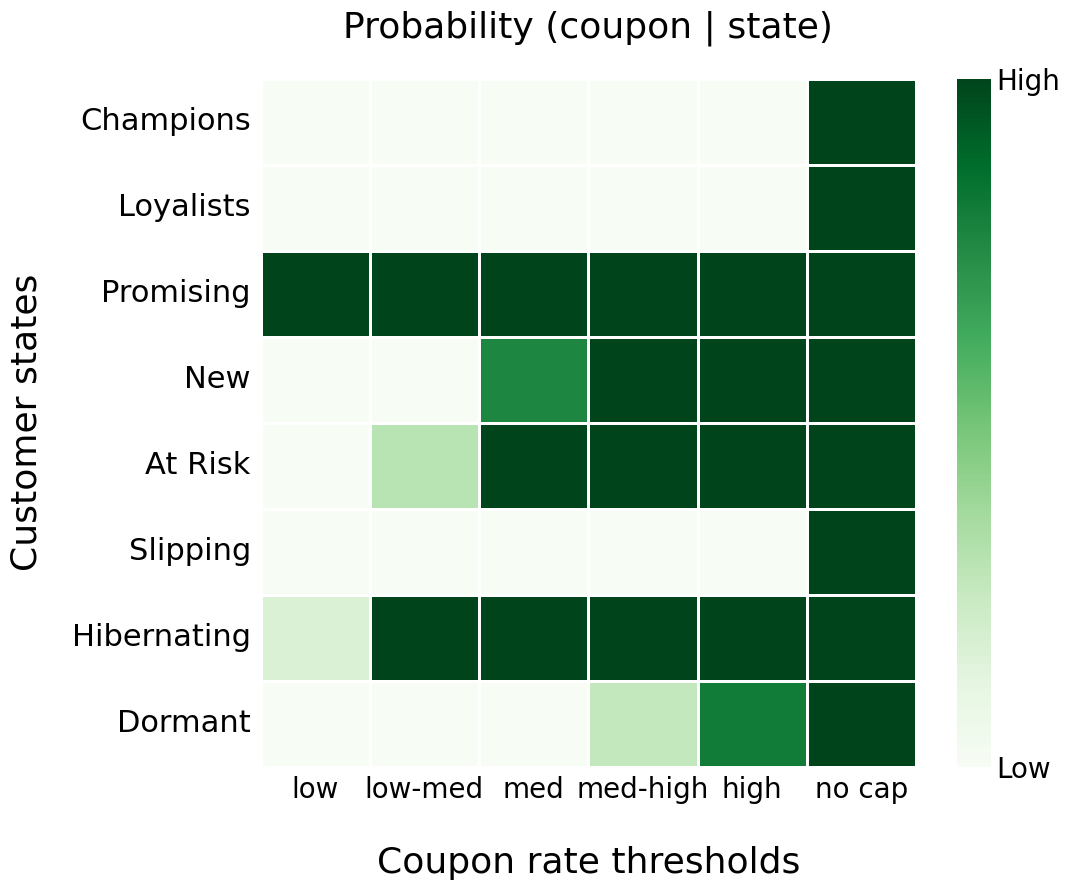}
    \Description{Optimal policies under coupon caps}
    \captionsetup{skip=2pt}
    \caption{Optimal policies under coupon caps}
    \label{fig:opt_policies}
\end{figure}

We benchmark the (a) \textit{Optimal Policy} ($\pi^\star$) obtained from solving the LP against the (b) \textit{Behavior Policy} ($\hat{\pi}_b$), i.e., the historical coupon allocation pattern across RFM segments, and the (c) \textit{Myopic Policy}, which greedily maximizes the immediate reward $\hat{R}(s,a)$. All three are obtained and evaluated offline under the observed coupon rate cap. While the behavior policy distributes relatively more coupons to champions and loyalists, the optimal policy redirects some of these coupons toward lapsed high-value customers (at risk, hibernating) and recent low-frequency customers (promising, new) who need the incentive to return or form habits (see Figure~\ref{fig:opt_policies}). The optimal and myopic policies diverge most notably on dormant versus new customers. While the myopic policy prioritizes recently dormant customers for their larger immediate uplift, the optimal policy invests in new customers whose habit formation compounds into higher lifetime value. Under offline evaluation, with causal dynamics estimated on a temporal holdout, the optimal policy yields an estimated 19\% and 8\% improvement in expected customer LTV over the behavior and myopic baselines, respectively. The offline experiments could help select a winning candidate for online validation.

\subsection{Coupon Budget Planning}

A key strength of the LP formulation is its ability to model budget constraints and business guardrails. Figure~\ref{fig:opt_policies} visualizes the optimal policy $\pi^\star(a \mid s)$ as the coupon budget scales from highly restrictive to unconstrained. At the historical coupon rate, the optimal policy is highly selective regarding which rider segments receive coupons. However, when the budget constraint is removed, coupons are distributed across all stages of the customer lifecycle. Notably, the \textit{Promising} and \textit{Champion} segments yield the highest state values or expected LTV.

\begin{figure}[h]
    \centering
    \includegraphics[width=0.45\textwidth]{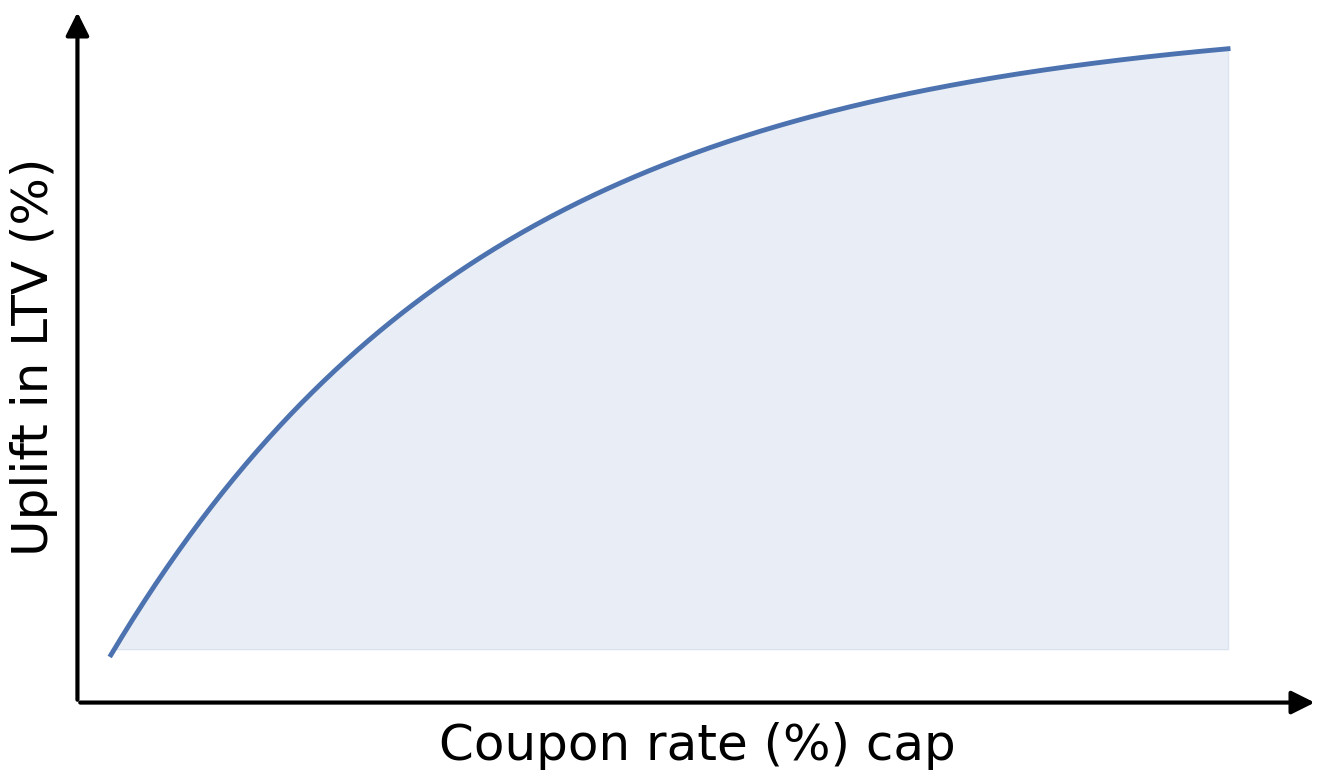}
    \Description{Fitted uplift in LTV under coupon caps}
    \captionsetup{skip=2pt}
    \caption{Fitted uplift in LTV under coupon caps}
    \label{fig:budget_planning}
\end{figure}

Beyond policy optimization, the LP formulation provides a practical tool for budget planning and scenario analysis. We evaluate returns across varying budget levels and fit an exponential saturation function to trace the budget-performance trade-off curve in Figure~\ref{fig:budget_planning}. This curve shows diminishing marginal returns on coupon expenditures, helping managers identify effective  spending thresholds.

\section{Conclusion and Future Work}

We propose a general-purpose, auditable action-selection module for business action targeting. The NBA framework unifies causal inference, counterfactual evaluation, model-based offline RL, and constrained optimization to maximize long-term value under six classes of operational constraints. We validate the framework on a coupon targeting case study in rideshare. Under offline evaluation, with causal dynamics estimated on a temporal holdout, the optimal policy provides an estimated 19\% and 8\% improvement in expected customer LTV over the behavior and myopic baselines, respectively. The framework additionally supports robust budget planning and what-if scenario analysis.

The results are based on offline policy evaluation and can help identify a winning candidate policy for potential online validation. Specifically, an online A/B test against the behavior policy may be conducted to further validate these findings. Future work could expand the state representation and incorporate multiple actions to test the framework's action-agnostic design. However, if we generalize to richer states and multi-actions, sparse data coverage may arise, requiring pessimism to be modeled explicitly \cite{kumar2020conservative, levine2020offline, jeunen2021pessimistic}. Finally, additional static user features could be modeled as context, similar to contextual MDPs \cite{hallak2015contextual}, allowing for personalized transition and reward dynamics without exploding the state space.

\bibliographystyle{ACM-Reference-Format}
\bibliography{references}

\end{document}